\documentclass[final]{colt2026} 

\usepackage{times}
\usepackage{amsmath,amssymb,mathtools}
\usepackage{bm}
\usepackage{booktabs}
\usepackage{multirow}
\usepackage{algorithmic}
\SetAlCapHSkip{0pt}
\SetAlgoCaptionSeparator{}
\renewcommand{\algocfconts}[3]{%
  \hrule height.8pt depth0pt\kern2pt
  #2\label{#1}\par
  \kern2pt\hrule height.4pt depth0pt\kern2pt
  #3\par
  \kern2pt\hrule height.4pt depth0pt
}

\newcommand{\R}{\mathbb{R}}
\newcommand{\E}{\mathbb{E}}
\newcommand{\norm}[1]{\left\|#1\right\|}
\newcommand{\Frob}[1]{\left\|#1\right\|_{F}}
\newcommand{\inner}[2]{\left\langle #1,\,#2\right\rangle}

\newcommand{\K}{\mathcal{K}}
\newcommand{\N}{\mathcal{N}}
\newcommand{\C}{\mathcal{C}}

\newcommand{\FCC}{\mathcal{C}_{\mathtt{FCC}}}

\newcommand{\vv}{\mathbf{v}}
\newcommand{\vx}{\mathbf{x}}
\newcommand{\q}{\mathbf{q}}
\newcommand{\vy}{\mathbf{y}}
\newcommand{\vz}{\mathbf{z}}
\newcommand{\vd}{\mathbf{d}}
\newcommand{\w}{\mathbf{w}}
\newcommand{\vq}{\mathbf{q}}
\newcommand{\vc}{\mathbf{c}}

\newcommand{\vzero}{\mathbf{0}}
\newtheorem{assumption}{Assumption}
\DeclareMathOperator*{\argmin}{arg\,min}
\newcommand{\LineComment}[1]{\hfill$\rhd\ $\text{#1}}
\title[Optimal Regret for Communication-Efficient Decentralized Online Learning]{%
Optimal Regret for Communication-Efficient Decentralized Online Learning}
\coltauthor{%
  \Name{Sifan Yang} \Email{yangsf@lamda.nju.edu.cn}\\
  \Name{Wenhao Yang} \Email{yangwh@lamda.nju.edu.cn}\\
  \addr State Key Laboratory of Novel Software Technology, Nanjing University, Nanjing 210023, China\\
  \addr School of Artificial Intelligence, Nanjing University, Nanjing 210023, China
  \AND
    \Name{Wei Jiang} \Email{jiang\_wei@njust.edu.cn}\\
    \addr School of Computer Science and Engineering, Nanjing University of Science and Technology, Nanjing, China
    \AND
  \Name{Yuanyu Wan} \Email{wanyy@zju.edu.cn}\\
  \addr State Key Laboratory of Blockchain and Data Security, Zhejiang University, Hangzhou, China\\
  \addr School of Software Technology, Zhejiang University, Ningbo, China
  \AND
    \Name{Lijun Zhang} \Email{zhanglj@lamda.nju.edu.cn}\\
  \addr State Key Laboratory of Novel Software Technology, Nanjing University, Nanjing 210023, China\\
  \addr School of Artificial Intelligence, Nanjing University, Nanjing 210023, China
}
\begin{document}
\maketitle
\thispagestyle{plain}

\begin{abstract}
In this paper, we investigate decentralized online convex optimization with compressed communication, where $n$ learners over a network jointly minimize a sequence of global loss functions based only on their local losses and compressed messages from their neighbors. The best known regret bounds for this problem are $O(\omega^{-1/2}n\rho^{-1}\sqrt{T\ln n})$ and $O(\omega^{-1}n\rho^{-2}\ln n\ln T)$ for convex and strongly convex loss functions, respectively, where $\rho\in(0,1)$ is the spectral gap of the communication matrix and $\omega\in(0,1]$ is the compression quality factor. Compared with the corresponding lower bounds of $\Omega(\omega^{-1/2}n\rho^{-1/4}\sqrt{T})$ and $\Omega(\omega^{-1}n\rho^{-1/2}\ln T)$, these upper bounds exhibit polynomial gaps in their dependence on $\rho^{-1}$. To close these gaps, we first develop Ascent-gossip, which combines an online compression strategy with a Chebyshev recursion and attains a faster consensus rate than existing compressed gossip methods. Furthermore, by adopting Ascent-gossip, we propose an online algorithm that achieves optimal regret bounds of $\widetilde O(\omega^{-1/2}n\rho^{-1/4}\sqrt{T})$ and $\widetilde O(\omega^{-1}n\rho^{-1/2}\ln T)$ for convex and strongly convex loss functions, respectively. Up to logarithmic factors in $n$ and $\rho^{-1}$, these results match the lower bounds in terms of $T$, $n$, $\rho$, and $\omega$.
\end{abstract}

\section{Introduction}
Decentralized online convex optimization (D-OCO) \citep{yan2012distributed,hosseini2013online,zhang2017projection,wan2024nearly,wan2025black,yang2026distributed,zhou2026} is a fundamental model for distributed learning with streaming data. Formally, D-OCO proceeds as a repeated game between an adversary and $n$ local learners connected by an undirected graph $\mathcal{G}=([n],E)$ with edge set $E\subseteq[n]\times[n]$. In each round $t\in[T]$, every learner $i\in[n]$ selects a decision $\w_i(t)$ from a convex set $\K\subseteq\R^d$. The adversary chooses convex loss functions $f_{t,1}(\cdot),\ldots,f_{t,n}(\cdot)$ on $\K$, and learner $i$ observes only its local function $f_{t,i}(\cdot)$. The performance of learner $i$ is measured by the \emph{global regret}
\begin{equation}\label{regret}
    R(T,i)=\sum_{t=1}^T\sum_{j=1}^n f_{t,j}(\w_i(t))-\min_{\w\in\K}\sum_{t=1}^T\sum_{j=1}^n f_{t,j}(\w),
\end{equation}
which compares the cumulative loss of its decisions under the global function $f_t(\cdot)=\sum_{j=1}^n f_{t,j}(\cdot)$ with that of the best fixed decision in hindsight. \citet{wan2024nearly} establish optimal regret bounds of $O(\rho^{-1/4}n\sqrt{T\ln n})$ and $O(\rho^{-1/2}n\ln n\ln T)$ for convex and strongly convex loss functions, respectively, where $\rho\in(0,1)$ is the spectral gap of the communication matrix $P$.

To acquire information about the global functions, each learner  needs to exchange messages with its neighbors, and the  communication overhead may be substantial. Therefore, several studies \citep{tu2022distributed,CAO2023111186,yang2026distributed,zhou2026} compress the messages to reduce the number of transmitted bits. \citet{tu2022distributed} and \citet{CAO2023111186} establish the regret bounds of $O(n\max\{\omega^{-2}\rho^{-4}n^{1/2},\omega^{-4}\rho^{-8}\}\sqrt{T})$ and $O(n\max\{\omega^{-2}\rho^{-4}n^{1/2},\omega^{-4}\rho^{-8}\}\ln T)$  for convex and strongly convex loss functions, respectively. These results exhibit a high dependence on the compression quality factor $\omega\in(0,1]$ and the spectral gap $\rho$, as well as a super-linear dependence on the number of learners $n$. Recently, \citet{yang2026distributed}  improve the regret bounds to $O(\omega^{-1/2}n\rho^{-1}\sqrt{T\ln n})$ and $O(\omega^{-1}n\rho^{-2}\ln n\ln T)$ for convex and strongly convex loss functions, respectively. They further establish the corresponding lower bounds of $\Omega(\omega^{-1/2}n\rho^{-1/4}\sqrt{T})$ and $\Omega(\omega^{-1}n\rho^{-1/2}\ln T)$. Subsequently, \citet{zhou2026} develop a simpler algorithm that achieves regret bounds of the same order.  Despite these advances, a large gap remains between the existing upper and lower bounds.  This raises a fundamental question \emph{whether we can achieve optimal regret bounds for D-OCO with compressed communication}?

\begin{table*}[t]
\centering
\small
\renewcommand{\arraystretch}{2}
\setlength{\tabcolsep}{5pt}
\caption{Comparison of existing regret bounds for D-OCO with compressed communication. Here, $n$ is the number of learners, $\rho<1$ is the spectral gap of the communication matrix, and $\omega\in(0,1]$ is the compression parameter. The regret gap denotes the multiplicative factor between each upper bound and the corresponding lower bound.}
\label{tab:regret-comparison}
\begin{tabular}{@{}cccc@{}}
\toprule[1pt]
\textbf{Source} & \textbf{Loss functions} & \textbf{Regret bounds} & \textbf{Regret gap} \\
\midrule
\multirow{2}{*}{\shortstack[c]{\textbf{Lower bounds}\\\citep{yang2026distributed}}}
& Convex
& $\Omega\!\left(\omega^{-1/2}\rho^{-1/4}n\sqrt{T}\right)$
& $-$ \\
& Strongly convex
& $\Omega\!\left(\omega^{-1}\rho^{-1/2}n\ln T\right)$
& $-$ \\
\midrule
\midrule
\multirow{2}{*}{\shortstack[c]{\citet{yang2026distributed}\\\citet{zhou2026}}}
& Convex
& $O\!\left(\omega^{-1/2}\rho^{-1}n\sqrt{\ln n}\sqrt{T}\right)$
& $O\!\left(\rho^{-3/4}\sqrt{\ln n}\right)$ \\
& Strongly convex
& $O\!\left(\omega^{-1}\rho^{-2}n\ln n\ln T\right)$
& $O\!\left(\rho^{-3/2}\ln n\right)$ \\
\midrule
\multirow{2}{*}{\textbf{This work}}
& Convex
& $O\!\left(\omega^{-1/2}\rho^{-1/4}n\sqrt{\ln n\ln(\rho^{-1})}\sqrt{T}\right)$
& $O\!\left(\sqrt{\ln n\ln(\rho^{-1})}\right)$ \\
& Strongly convex
& $O\!\left(\omega^{-1}\rho^{-1/2}n\ln n\ln(\rho^{-1})\ln T\right)$
& $O\!\left(\ln n\ln(\rho^{-1})\right)$ \\
\bottomrule[1pt]
\end{tabular}
\end{table*}

To close this gap, we first design a faster gossip strategy with compressed communication, named Ascent-gossip, and use it to develop an online algorithm with optimal regret bounds. The existing algorithms \citep{yang2026distributed,zhou2026} adopt Choco-gossip \citep{koloskova2019decentralized}, which builds on the standard gossip \citep{xiao2004fast} and suffers from a slow consensus rate. Consequently, these algorithms require a large number of communication rounds to reduce the consensus error, which leads to a suboptimal dependence of their regret bounds on $\rho$. In the absence of compression, accelerated gossip \citep{liu2011accelerated} attains faster consensus through the Chebyshev recursion. Thus, a natural approach is to incorporate the  Chebyshev recursion into Choco-gossip to accelerate the consensus rate. However, this combination fails to retain this acceleration under compressed communication.  The recursion amplifies the compression error introduced at each iteration through all subsequent ones, and this accumulation eliminates the advantage of the Chebyshev recursion over standard gossip. To address this amplification, we adopt fast compressed communication ($\mathtt{FCC}$) \citep{huang2022lower}, which repeatedly compresses the residual and reduces the compression error at each gossip iteration exponentially in the number of compressions. Our analysis reveals that the Chebyshev recursion retains its acceleration as long as the compression error at each step remains sufficiently small. Therefore, we propose Ascent-gossip, which integrates $\mathtt{FCC}$ into every step of the Chebyshev recursion and achieves accelerated consensus under compressed communication.  Ascent-gossip requires $O(\omega^{-1}\rho^{-1/2}\ln(1/\rho)\ln(1/\epsilon))$ communication rounds to reduce the consensus and compression errors to an $\epsilon$ fraction of their initial values, which improves upon the $O(\omega^{-1}\rho^{-2}\ln(1/\epsilon))$ rounds of Choco-gossip.

 Furthermore, we incorporate Ascent-gossip into the follow-the-regularized-leader (FTRL) framework \citep{hazan2016introduction} and develop Ascent-FTRL. Since each Ascent-gossip iteration requires multiple communication rounds, we adopt the blocking update mechanism \citep{wan2022projection,wan2024nearly,yang2026distributed}, which divides the $T$ rounds into blocks and distributes the required communications across the rounds within each block, so that each round involves only one communication. For convex and strongly convex loss functions, Ascent-FTRL achieves regret bounds of $\widetilde O(\omega^{-1/2}n\rho^{-1/4}\sqrt{T})$ and $\widetilde O(\omega^{-1}n\rho^{-1/2}\ln T)$, respectively. These bounds match the lower bounds of \citet{yang2026distributed} up to logarithmic factors in $n$ and $\rho^{-1}$, and close the gap in $\rho$. We compare our results with existing ones in Table~\ref{tab:regret-comparison}.

\section{Related Work}
In this section, we briefly review related work on decentralized online convex optimization and compressed communication.

\subsection{Decentralized Online Convex Optimization (D-OCO)}
D-OCO generalizes online convex optimization (OCO) \citep{hazan2016introduction} from a single learner to $n\ge 2$ learners connected by an undirected graph $\mathcal{G}=([n],E)$, with communication restricted to its edges. Each learner $i$ aims to minimize its regret with respect to the global loss function $f_t(\cdot)=\sum_{j=1}^n f_{t,j}(\cdot)$, while observing only its local loss function $f_{t,i}(\cdot)$. Information about ${f_{t,j}(\cdot)}_{j\neq i}$ is acquired through gossip with neighboring learners, which computes weighted averages of the exchanged variables according to a gossip matrix $P$. The pioneering work of \citet{yan2012distributed} introduces D-OGD, a decentralized variant of OGD \citep{zinkevich2003online}, which achieves regret bounds of $O(n^{5/4}\rho^{-1/2}\sqrt{T})$ and $O(n^{3/2}\rho^{-1}\ln T)$ for convex and strongly convex loss functions, respectively. D-FTRL \citep{hosseini2013online} instead applies standard gossip to the accumulated local gradients of FTRL \citep{shalev2007primal} and attains the same regret bound for convex loss functions. \citet{wan2024nearly} develop an online accelerated gossip strategy and improve the regret bounds to $O(n\rho^{-1/4}\sqrt{T\ln n})$ and $O(n\rho^{-1/2}\ln n\ln T)$ for convex and strongly convex loss functions, respectively. The corresponding lower bounds of $\Omega(n\rho^{-1/4}\sqrt{T})$ \citep{wan2024nearly} and $\Omega(n\rho^{-1/2}\ln T)$ \citep{wan2024optimal} confirm the optimality of these results up to logarithmic factors in $n$. \citet{wan2025black} investigates D-OCO in changing environments through black-box reductions for adaptive and dynamic regret minimization. 

Since each learner observes only its local loss function, information about the global loss must be acquired through communication with other learners, which can incur substantial communication overhead. Existing studies \citep{zhang2017projection,Li2021,wan2022projection,tu2022distributed,CAO2023111186,yuan2022distributed,zhang2023quantized} reduce this overhead mainly by decreasing either the number of communication rounds or the amount of data transmitted in each round. We focus on the latter in this paper, while studies that reduce the number of communication rounds are discussed in Appendix \ref{additional}.

\subsection{Compressed Communication}

To reduce communication overhead in distributed optimization, extensive studies employ compression operators to reduce the amount of information transmitted between local nodes \citep{seide20141,stich2018sparsified,bernstein2018signsgd,karimireddy2019error,Li2021,huang2022lower,tu2022distributed,CAO2023111186,yang2026distributed,zhou2026}. Existing compression operators are commonly characterized by two properties, namely contractiveness and unbiasedness. A contractive compressor $\mathcal{C}:\mathbb{R}^d\rightarrow\mathbb{R}^d$ satisfies $\mathbb{E}_{\mathcal{C}}\left[\|\mathcal{C}(\mathbf{x})-\mathbf{x}\|^2\right]
\leq (1-\omega)\|\mathbf{x}\|^2$ for some $\omega\in(0,1]$, with representative examples including unscaled Rand-$k$ and Top-$k$ sparsification. An unbiased compressor satisfies $\mathbb{E}_{\mathcal{C}}[\mathcal{C}(\mathbf{x})]=\mathbf{x}$ and $\mathbb{E}_{\mathcal{C}}\left[\|\mathcal{C}(\mathbf{x})-\mathbf{x}\|^2\right]
\leq \nu\|\mathbf{x}\|^2$ for some $\nu\geq0$. Although unbiased compressors admit simpler theoretical analysis, biased compression schemes tend to outperform their unbiased counterparts in practice \citep{huang2022lower}.

In the field of D-OCO with compressed communication, \citet{Li2021} explore compressed decentralized online learning for near-sensor data analytics. Subsequently, \citet{tu2022distributed} and \citet{CAO2023111186}  propose a communication-efficient method by leveraging the compressed communication \citep{koloskova2019decentralized}, and establish $O(\max\{\omega^{-2}\rho^{-4}n^{1/2},\omega^{-4}\rho^{-8}\}n\sqrt{T})$ and $O(\max\{\omega^{-2}\rho^{-4}n^{1/2},\omega^{-4}\rho^{-8}\}n\ln{T})$ regret bounds for convex and strongly convex loss functions, respectively. Recently, \citet{yang2026distributed} develop Top-DOGD using a two-level blocking update mechanism, improving the regret bounds to $O(\omega^{-1/2}n\rho^{-1}\sqrt{T\ln n})$ and $O(\omega^{-1}n\rho^{-2}\ln n\ln T)$ for convex and strongly convex loss functions, respectively. They further provide the lower bounds of $\Omega(\omega^{-1/2}n\rho^{-1/4}\sqrt{T})$ and $\Omega(\omega^{-1}n\rho^{-1/2}\ln{T})$ for convex and strongly convex functions, respectively. Later, \citet{zhou2026} develop an algorithm that applies Choco-gossip to the accumulated gradients. Their algorithm attains the same regret bounds as Top-DOGD in the full-information setting, while improving both regret bounds and communication complexity in the bandit setting. Despite these advances, the existing regret bounds still exhibit polynomial gaps in their dependence on $\rho$ compared with the corresponding lower bounds.

\subsection{Gossip with Compressed Communication}

To reduce communication costs in offline decentralized optimization, several works incorporate compression into gossip algorithms. compression of the exchanged states can violate
average preservation, preventing average consensus.  To overcome this limitation, \citet{koloskova2019decentralized} propose Choco-gossip by using the difference compression technique \citep{tang2018communication}. Choco-gossip guarantees convergence under compressed communication, but its communication complexity has a quadratic dependence on $\rho^{-1}$. Later, \citet{Liao2021CGT} combine compression with gradient tracking and establishes linear convergence for smooth and strongly convex objectives. \citet{Zhao2022BEER} further develop compressed gradient tracking for smooth nonconvex optimization.

More closely to our work, SCG \citep{Toghani2022SCG} combines difference compression with momentum to accelerate average consensus. Its analysis relies on Metropolis--Hastings weights and characterizes the convergence rate through the network size rather than the spectral gap of a general prescribed gossip matrix. Moreover, its accelerated convergence guarantee with a constant consensus step size requires $\sqrt{1-\omega}=O(n^{-2})$, imposing a restrictive assumption on the compression quality as the network size increases.

\section{Preliminaries}\label{sec:prelim}

In this section, we present the notation and assumptions used in this paper. 

\paragraph{Notation.} We write $[n]=\{1,\ldots,n\}$, and $\mathbf{1}\in\R^n$ and $I_n$ denote the all-ones vector and the identity in $\R^{n\times n}$, respectively. For vectors, $\norm{\cdot}$ denotes the Euclidean norm. For matrices, $\norm{\cdot}_2$ denotes the spectral norm, while $\Frob{\cdot}$ is the Frobenius norm and $\inner{\cdot}{\cdot}$ is the Euclidean inner product. Given $\vx_1,\ldots,\vx_n\in\R^d$, we stack them as $X=[\vx_1,\ldots,\vx_n]\in\R^{d\times n}$ and set $J=\mathbf{1}\mathbf{1}^\top/n$, so every column of $XJ$ equals the average $\frac{1}{n}\sum_{i=1}^n\vx_i$. Finally, $\widetilde O(\cdot)$ hides constants and polylogarithmic factors in $n$ and $\rho$. Throughout the analysis, we consider $n\ge2$ learners.
For simplicity, we assume that $T$ is divisible by the
block size $L$, so that $B=T/L$ is an integer.
The general case can be handled by bounding the regret
of the final incomplete block separately.

Our analysis relies on the following assumptions, which are standard in D-OCO \citep{yan2012distributed,hosseini2013online,wan2024nearly,wan2024optimal}.

\begin{assumption}[Gossip matrix]\label{ass:matrix}
\textnormal{The  local learners communicate over an undirected graph $\mathcal{G}=([n],E)$ through a gossip matrix $P\in\R^{n\times n}$ that is fixed in advance rather than chosen by the algorithm. It is symmetric, doubly stochastic, and positive semidefinite, and its entries satisfy $P_{ij}>0$ only if $(i,j)\in E$ or $i=j$. Moreover, its second largest eigenvalue obeys $\lambda_2(P)<1$, so the spectral gap $\rho=1-\lambda_2(P)$ lies in $(0,1)$.}
\end{assumption}
\textbf{Remark.} Assumption~\ref{ass:matrix} implies $P_{ij}=0$ whenever $j\notin\N_i$ and $\sum_{j\in\N_i}P_{ij}=1$ for every $i\in[n]$, where $\N_i=\{j\in[n]\mid (i,j)\in E\}\cup\{i\}$ consists of learner $i$ and its neighbors. Since $P$ is symmetric and doubly stochastic, it also satisfies $PJ=JP=J$, which implies $(P-I)J=0$. Because $P$ is positive semidefinite, every eigenvalue of $P$ associated with an eigenvector orthogonal to $\mathbf{1}$ lies in $[0,1-\rho]$ and consequently $\norm{I-P}_2\le1$.

\begin{assumption}[Convexity]\label{ass:cvx}
\textnormal{For every $t\in[T]$ and $i\in[n]$, the local loss $f_{t,i}(\cdot)$ is convex over $\K$.}
\end{assumption}

\begin{assumption}[Strong convexity]\label{ass:scvx}
\textnormal{For some $\alpha>0$, the local loss $f_{t,i}(\cdot)$ is $\alpha$-strongly convex over $\K$, that is,
\begin{equation*}
    f_{t,i}(\vy)\ge f_{t,i}(\vx)+\inner{\nabla f_{t,i}(\vx)}{\vy-\vx}+\frac{\alpha}{2}\norm{\vy-\vx}^2,\qquad \forall\,\vx,\vy\in\K.
\end{equation*}}
\end{assumption}

\begin{assumption}[Bounded gradient norm]\label{ass:grad}
\textnormal{For every $t\in[T]$ and $i\in[n]$, the gradient of $f_{t,i}(\cdot)$ satisfies $\norm{\nabla f_{t,i}(\w)}\le G$ for all $\w\in\K$.}
\end{assumption}

\begin{assumption}[Bounded domain]\label{ass:domain}
\textnormal{The feasible set $\K\subseteq\R^d$ is convex, contains the origin $\vzero$, and satisfies $\norm{\w}\le D$ for all $\w\in\K$.}
\end{assumption}

\begin{assumption}[Compressor]\label{ass:compressor}
\textnormal{For some $\omega\in(0,1]$, the compressor $\C(\cdot)\colon\R^d\to\R^d$ satisfies
\begin{equation*}
    \E_{\C}\bigl[\norm{\C(\vv)-\vv}^2\bigr]\le(1-\omega)\norm{\vv}^2,\qquad \forall\,\vv\in\R^d,
\end{equation*}
where $\E_{\C}[\cdot]$ denotes the expectation over the internal randomness of $\C(\cdot)$ and $\omega =1$ means no compression.}
\end{assumption}

To mitigate the error of the standard compressors, \citet{huang2022lower} design the \textit{fast compressed communication} ($\mathtt{FCC}$), as summarized in  Algorithm~\ref{alg:fcc}. The core idea is to repeatedly apply the standard compressor for $M$ rounds and transmit the compressed data at each round, which involves $M$ rounds of communication. When $M=1$, it degenerates to the standard compressor. We state the following lemma from  \citet{huang2022lower} to establish the guarantee of $\FCC(\cdot, M)$.

\begin{algorithm}[t]
\floatconts{alg:fcc}{\caption{$\FCC(\vv,M)$}}{%
\begin{algorithmic}[1]
\STATE \textbf{Input:} vector $\vv\in\R^d$, $M\ge1$
\STATE Initialize $\vc^0=\vzero$
\FOR{$m=0,\ldots,M-1$}
\STATE Compute $\vq^m=\mathcal C(\vv-\vc^m)$ and transmit $\vq^m$
\STATE Update both accumulators as $\vc^{m+1}=\vc^m+\vq^m$
\ENDFOR
\end{algorithmic}
}
\end{algorithm}

\begin{lemma}[Lemma 2 in \citet{huang2022lower}]
\label{lem:fcc}\textnormal{
Under Assumption~\ref{ass:compressor}, applying $\FCC(\cdot, M)$ to a vector $\vv$ gives
\begin{equation*}
    \E\bigl[\norm{\FCC(\vv, M)-\vv}^2\bigr]\le(1-\omega)^M\norm{\vv}^2.
\end{equation*}}
\end{lemma}

\paragraph{Choco-gossip.}
Choco-gossip \citep{koloskova2019decentralized} is a classical gossip algorithm for decentralized optimization with compressed communication. It employs difference compression \citep{tang2018communication} to control the accumulation of compression errors and ensure consensus among local states. Each learner $i$ maintains auxiliary variables $\hat{\vx}_j(t)\in\mathbb{R}^d$ to record information received from its neighbors $j$, while $\hat{\vx}_i(t)\in\mathbb{R}^d$ tracks the information transmitted by learner $i$ in previous rounds. At each round $t$, learner $i$ computes
\begin{equation}\label{cho:update}
   \begin{aligned}
       \vx_i(t+1)&= \vx_i(t)+\gamma\sum_{j\in\mathcal{N}_i}P_{ij}(\hat{\vx}_j(t)-\hat{\vx}_i(t))\\
       \hat{\vx}_j(t+1)&=\hat{\vx}_j(t)+\C(\vx_j(t+1)-\hat{\vx}_j(t)), \forall j\in \N_i,
   \end{aligned} 
\end{equation} 
where $\gamma\in(0,1]$ is the consensus step size. Since $P$ is symmetric and doubly stochastic, Choco-gossip preserves the average of the iterates over all iterations, i.e., $\bar{\vx}(t+1)=\bar{\vx}(t)$, where $\bar{\vx}(t)=\frac{1}{n}\sum_{i=1}^n\vx_i(t)$. The following lemma establishes its convergence guarantee.

\begin{lemma}[Theorem 2 in \cite{koloskova2019decentralized}]\label{lem:k19} \textnormal{We define $e_t =  \E_{\C}[\sum_{i=1}^n \norm{\vx_i(t)-\bar{\vx}(t)}^2+\norm{\vx_i(t)-\hat{\vx}_i(t)}^2]$ and $\bar{\vx}(t)=\frac{1}{n}\sum_{i=1}^n\vx_i(t)$. Given an $\omega$-contractive compressor $\C(\cdot)$, for any round  $t$, by setting $\gamma = \frac{\rho\omega}{16\rho+\rho^2+4\beta^2+2\rho\beta^2-8\rho\omega}$, $\beta = \norm{I-P}_2$ and performing the update in (\ref{cho:update}) for $K$ rounds, we can ensure 
\begin{equation*}
       e_{t+K} \leq \left(1-\frac{\rho^2\omega}{82}\right)^{K} e_t.
    \end{equation*}}
\end{lemma}
\paragraph{Remark.}
Lemma~\ref{lem:k19} shows that Choco-gossip can effectively reduce the consensus and compression errors, ensuring convergence under compressed communication. Nevertheless, reducing the joint error from $e_t$ to $\epsilon e_t$ requires $O(\omega^{-1}\rho^{-2}\ln(1/\epsilon))$ communication rounds.

\section{Main Results}
\label{sec:main}

In this section, we first present our accelerated gossip method and establish its convergence guarantee. We then utilize it to develop an online algorithm that achieves the optimal regret bounds.

\subsection{Accelerated Gossip with Compressed Communication}
\label{sec:acg}


Without compressed communication, accelerated gossip \citep{liu2011accelerated} employs the Chebyshev recursion to achieve faster consensus than standard gossip \citep{xiao2004fast}. Specifically, this recursion incorporates the previous state into the current gossip update. With $\vx_i^{(-1)}=\vx_i^{(0)}$, each learner $i$ updates its local state according to
\begin{equation*}
\vx_i^{(k+1)}=(1+\theta)\sum_{j\in\mathcal{N}_i}P_{ij}\vx_j^{(k)}-\theta\vx_i^{(k-1)},
\end{equation*}
where $\theta$ is the mixing coefficient.  The standard gossip step corresponds to $\theta=0$ and requires $O(\rho^{-1}\ln(1/\epsilon))$ iterations to reduce the consensus error by a factor of $\epsilon$, whereas the Chebyshev recursion reduces this requirement to $O(\rho^{-1/2}\ln(1/\epsilon))$.

A natural approach to accelerating consensus under compressed communication is to apply the Chebyshev recursion to the compressed gossip update of Choco-gossip. However, a direct combination does not guarantee acceleration, since the recursion amplifies the compression error introduced at each iteration. Our analysis of the recursion under compression reveals that the acceleration is retained as long as the relative compression error at each iteration is reduced to $O(\rho^2)$. $\mathtt{FCC}$ \citep{huang2022lower} meets this requirement at a logarithmic communication cost.  According to Lemma~\ref{lem:fcc}, setting the number of compression rounds to $M=\lceil 2\omega^{-1}\ln(1/\rho)\rceil$ ensures that, for any vector $\mathbf{x}$,
\begin{equation*}
    \mathbb{E}\left[\left\|\FCC(\mathbf{x},M)-\mathbf{x}\right\|^2\right]\le\rho^2\|\mathbf{x}\|^2.
\end{equation*}

Therefore, we replace the standard compressor in Choco-gossip with $\mathtt{FCC}$ and integrate the resulting update into the Chebyshev recursion, which yields \underline{a}ccelerated go\underline{s}sip with \underline{c}ompr\underline{e}ssed commu\underline{n}ica\underline{t}ion (Ascent-gossip), summarized in Algorithm~\ref{alg:acg}. Each learner $i$ initializes $\mathbf{x}_i^{(-1)}=\mathbf{x}_i^{(0)}$ and maintains a shared estimate $\hat{\mathbf{x}}_j^{(k)}$ for every $j\in\mathcal{N}_i$, which serves as a compressed replica of $\mathbf{x}_j^{(k)}$.  At iteration $k$, learner $i$ applies $\mathtt{FCC}$ to the difference between its current state and its shared estimate, i.e.,
\begin{equation*}
\mathbf{q}_i^{(k)}=\FCC\!\left(\mathbf{x}_i^{(k)}-\hat{\mathbf{x}}_i^{(k)},M\right).
\end{equation*}
Each $\mathtt{FCC}$ consists of $M$ rounds of compressed communication, during which learner $i$ sends the compressed residuals that constitute $\mathbf{q}_i^{(k)}$ to its neighbors and receives those of $\mathbf{q}_j^{(k)}$ from each $j\in\mathcal{N}_i$. After these $M$ rounds, learner $i$ updates the shared estimates as
\begin{equation*}
\hat{\mathbf{x}}_j^{(k+1)}=\hat{\mathbf{x}}_j^{(k)}+\mathbf{q}_j^{(k)},\quad\forall j\in\mathcal{N}_i.
\end{equation*}
With the updated estimates, learner $i$ then performs the accelerated gossip update
\begin{equation*}
\mathbf{x}_i^{(k+1)}=(1+\theta)\left(\mathbf{x}_i^{(k)}+\sum_{j\in\mathcal{N}_i}P_{ij}\bigl(\hat{\mathbf{x}}_j^{(k+1)}-\hat{\mathbf{x}}_i^{(k+1)}\bigr)\right)-\theta\mathbf{x}_i^{(k-1)}.
\end{equation*}
Note that if the communication is exact, i.e., $\omega=1$ and $M=1$, we have $\hat{\mathbf{x}}_j^{(k+1)}=\mathbf{x}_j^{(k)}$ for all $j\in\mathcal{N}_i$, and the above update reduces to accelerated gossip \citep{liu2011accelerated}.

\begin{algorithm}[t]
\floatconts{alg:acg}{\caption{Ascent-gossip}}{%
\begin{algorithmic}[1]
\STATE \textbf{Input:} $\{\vx_i^{(0)}\}_{i\in[n]}$, $\{\hat\vx_i^{(0)}\}_{i\in[n]}$, $P$, $\theta$, and $K,M$
\STATE Set $\vx_i^{(-1)}=\vx_i^{(0)}$ for all $i\in[n]$
\FOR{$k=0,\ldots,K-1$ \textbf{in parallel for each} $i\in[n]$}
\STATE Compress $\vq_i^{(k)}=\FCC(\vx_i^{(k)}-\hat\vx_i^{(k)},M)$
\STATE Send $\vq_i^{(k)}$ to the neighbors $\N_i$ and obtain $\vq_j^{(k)}$ from $j\in\mathcal{N}_i$ \\ \LineComment{\textcolor{blue}{Require $M$ communication rounds}}
\STATE Update $\hat\vx_j^{(k+1)}=\hat\vx_j^{(k)}+\vq_j^{(k)}$ for $j\in\mathcal{N}_i\cup\{i\}$
\STATE $\vx_i^{(k+1)}=(1+\theta)\left(\vx_i^{(k)}+\sum_{j\in\mathcal{N}_i}P_{ij}(\hat\vx_j^{(k+1)}-\hat\vx_i^{(k+1)})\right)-\theta\vx_i^{(k-1)}$
\ENDFOR
\end{algorithmic}
}
\end{algorithm}

We next establish the convergence guarantee of Ascent-gossip. 

\begin{theorem}
\label{thm:acg}
\textnormal{Let $\bar{\vx}=\frac1n\sum_{i=1}^n\vx_i^{(0)}$.
We define the sum of consensus and compression errors as $e_t=\sum_{i=1}^n \left( \norm{\vx_i^{(t)}-\bar{\vx}}^2+\norm{\vx_i^{(t)}-\hat{\vx}_i^{(t)}}^2 \right).$ Under Assumptions~\ref{ass:matrix} and~\ref{ass:compressor}, by setting $M = \lceil2\omega^{-1}\ln\left(49\rho^{-1}\right)\rceil$ and  $\theta=\frac{1}{1+\sqrt{2\rho-\rho^2}}$,  Algorithm~\ref{alg:acg} ensures
\begin{equation*}
    \E[e_t]\le (3+\sqrt5)\left(1-\frac{\sqrt\rho}{4}\right)^t\E[e_0].
\end{equation*}
}
\end{theorem}

\paragraph{Remark.}
A key distinction between Ascent-gossip and Choco-gossip is that Ascent-gossip does not require an additional consensus stepsize $\gamma$. Choco-gossip introduces $\gamma$ to control the effect of compression errors on consensus updates. In contrast, Ascent-gossip reduces the compression error to $O(\rho^2)$ through $\mathtt{FCC}$, thereby eliminating the need for an additional consensus stepsize. Another distinction arises under exact communication. With the prescribed stepsize $\gamma<1$, Choco-gossip does not reduce to standard gossip, whereas Ascent-gossip with $\theta=0$ reduces to $\mathbf{X}^{k+1}=\mathbf{X}^kP$, which is precisely the standard gossip update.

\paragraph{Remark.} Under exact communication, the states satisfy $\hat{\mathbf{x}}_i^{(k+1)}=\mathbf{x}_i^{(k)}$, and Ascent-gossip reduces to accelerated gossip \citep{liu2011accelerated}. Moreover, Theorem~\ref{thm:acg} applies to any $\omega$-contractive compressor satisfying Assumption~\ref{ass:compressor} and does not rely on a particular compression scheme.

\paragraph{Remark.} One might notice that each Ascent-gossip iteration requires $M=\Theta(\omega^{-1}\ln\rho^{-1})$ communication rounds due to $\mathtt{FCC}$, raising the concern that the acceleration may simply result from the additional communication.  We compare Ascent-gossip and Choco-gossip under the same target accuracy. To ensure $\E[e_t]\leq\epsilon\E[e_0]$, Ascent-gossip requires $O(\omega^{-1}\rho^{-1/2}\ln\rho^{-1}\ln(1/\epsilon))$ communication rounds, whereas Choco-gossip \citep{koloskova2019decentralized}
 requires $O(\omega^{-1}\rho^{-2}\ln(1/\epsilon))$ communication rounds.  Therefore, even after accounting for the additional communication required by $\mathtt{FCC}$, Ascent-gossip achieves a substantially lower communication complexity than Choco-gossip and converges faster under the same communication budget.

\subsection{Ascent-FTRL}
After establishing the convergence guarantee of Ascent-gossip, we further incorporate it into D-OCO. Inspired by \citet{zhou2026}, we develop Ascent-FTRL by applying Ascent-gossip to the cumulative gradient states in the FTRL update, as summarized in Algorithm~\ref{alg:acftrl}.

During block $b$, learner $i$ plays the fixed decision $\w_i(b)$ and accumulates the local gradient information as $\vd_i(b)=\sum_{t=(b-1)L+1}^{bL}\bigl(\nabla f_{t,i}(\w_i(b))-\alpha\w_i(b)\bigr)$. At the beginning of block $b\geq2$, learner $i$ incorporates the gradient information from the previous block into its cumulative state by setting $\vz_i^{(0)}(b)=\vz_i(b-1)+\vd_i(b-1)$ and initializes the shared state as $\hat{\vz}_i^{(0)}(b)=\hat{\vz}_i(b-1)$. We further set $\vz_i^{(-1)}(b)=\vz_i^{(0)}(b)$ to initialize the Chebyshev iteration in Ascent-gossip.

Starting from $\vz_i^{(0)}(b)$ and $\hat{\vz}_i^{(0)}(b)$, the learners perform $K$ iterations of Ascent-gossip and set $\vz_i(b)=\vz_i^{(K)}(b)$ and $\hat{\vz}_i(b)=\hat{\vz}_i^{(K)}(b)$. Learner $i$ then updates its decision as

\begin{algorithm}[t]
\floatconts{alg:acftrl}{\caption{Ascent-FTRL}}{%
\begin{algorithmic}[1]
\STATE \textbf{Input:} $T$, $\eta>0$,  $\alpha\ge0$, $\theta$, $K,M$, and block size $L=KM$
\STATE Set $\w_i(1)=\w_i(2)=\vz_i(1)=\hat{\vz}_i(1)=\vzero$ for all $i\in[n]$
\FOR{$b=1,\ldots,T/L$ \textbf{in parallel for each} $i\in[n]$}
\FOR{$t=(b-1)L+1,\ldots,bL$}
\STATE Play $\w_i(b)$ and observe $\nabla f_{t,i}(\w_i(b))$
\ENDFOR
\IF{$b\geq 2$}
\STATE $\vz_i^{(0)}(b)=\vz_i(b-1)+\vd_i(b-1)$ and $\hat{\vz}_i^{(0)}(b)=\hat{\vz}_i(b-1)$
\STATE Set $\vz_i^{(-1)}(b)=\vz_i^{(0)}(b)$
\STATE Compute $\vz_i^K(b)$ and $\hat{\vz}_i^K(b)$ by invoking Algorithm~\ref{alg:acg} for $L$ rounds
\STATE $\vz_i(b)=\vz_i^{(K)}(b)$ and $\hat{\vz}_i(b)=\hat{\vz}_i^{(K)}(b)$
\ENDIF
\STATE $\vd_i(b)=\sum_{t=(b-1)L+1}^{bL}\bigl(\nabla f_{t,i}(\w_i(b))-\alpha\w_i(b)\bigr)$
\STATE If $2\le b<T/L$, compute $\w_i(b+1)$  according to \eqref{eq:aligned-local-update}
\ENDFOR
\end{algorithmic}
}
\end{algorithm}

\begin{equation}
\label{eq:aligned-local-update}
\w_i(b+1)=
\begin{cases}
\displaystyle\argmin_{\w\in\K}\left\{\inner{\vz_i(b)}{\w}+\frac{1}{\eta}\norm{\w}^2\right\}, & \alpha=0,\\
\displaystyle\argmin_{\w\in\K}\left\{\inner{\vz_i(b)}{\w}+\frac{(b-1)L\alpha}{2}\norm{\w}^2\right\}, & \alpha>0.
\end{cases}
\end{equation}

Let $\bar{\vz}(b)=\frac{1}{n}\sum_{i=1}^n\vz_i(b)$ be the consensus. We first establish the following approximation-error bound for Ascent-FTRL.
\begin{lemma}
\label{lem:consensus}
\textnormal{Under Assumptions~\ref{ass:matrix}, \ref{ass:grad}, \ref{ass:domain}, and~\ref{ass:compressor}, for every $i\in[n]$ and every state $\vz_i(b)$,  Algorithm \ref{alg:acftrl} with
\begin{equation}\label{eq:para}
    \theta = \frac{1}{1+\sqrt{2\rho-\rho^2}}, K = \left\lceil\frac{4\ln(24n)}{\sqrt{\rho}}\right\rceil, M =\lceil2\omega^{-1}\ln\left(49\rho^{-1}\right)\rceil, L=KM
\end{equation}
ensures 
\begin{equation*}
\E\bigl[\norm{\vz_i(b)-\bar\vz(b)}_2\bigr]\le L(G+\alpha D) = O\left(\omega^{-1}\rho^{-1/2}\ln{n}\ln{\rho^{-1}}\right).    
\end{equation*}}
\end{lemma}
\textbf{Remark.}
Existing methods \citep{zhou2026} only yield an $O(\omega^{-1}\rho^{-2}\ln n)$ bound with Choco-gossip. In contrast, Lemma~\ref{lem:consensus} achieves a better $O(\omega^{-1}\rho^{-1/2}\ln n\ln\rho^{-1})$ bound, owing to the improved consensus achieved by Ascent-gossip. By exploiting this improvement, we establish the following guarantees for Ascent-FTRL.

\begin{theorem}
\label{thm:cvx}
\textnormal{Under Assumptions~\ref{ass:matrix}, \ref{ass:cvx}, \ref{ass:grad}, \ref{ass:domain} and \ref{ass:compressor}, for any $i\in[n]$,  Algorithm~\ref{alg:acftrl} with $\theta, K, M, L$ defined in \eqref{eq:para} and $\eta = \frac{D}{G\sqrt{LT}}$ ensures
\begin{equation*}
    \E\left[R(T,i)\right]\leq O(\omega^{-1/2}n\rho^{-1/4}\sqrt{T\ln{n}\ln{(\rho^{-1})}}).
\end{equation*}}
\end{theorem}

\begin{theorem}
\label{thm:scvx}
\textnormal{Under Assumptions~\ref{ass:matrix}, \ref{ass:scvx}, \ref{ass:grad}, \ref{ass:domain} and \ref{ass:compressor}, for any $i\in[n]$,  Algorithm~\ref{alg:acftrl} with $\theta, K, M$ and $L$ defined in \eqref{eq:para} ensures
\begin{equation*}
    \E\left[R(T,i)\right]\leq O(\omega^{-1}n\rho^{-1/2}\ln{T}\ln{n}\ln{(\rho^{-1})}).
\end{equation*}}
\end{theorem}
\paragraph{Remark.}
On one hand, compared with the existing regret bounds of $O(\omega^{-1/2}\rho^{-1}n\sqrt{T\ln n})$ and $O(\omega^{-1}\rho^{-2}n\ln n\ln T)$ for convex and strongly convex loss functions \citep{yang2026distributed,zhou2026}, Ascent-FTRL improves the dependence on the spectral gap from $\rho^{-1}$ to $\rho^{-1/4}$ and from $\rho^{-2}$ to $\rho^{-1/2}$, respectively.  On the other hand, together with the lower bounds of $\Omega(\omega^{-1/2}\rho^{-1/4}n\sqrt{T})$ and $\Omega(\omega^{-1}\rho^{-1/2}n\ln T)$ established in \citet{yang2026distributed}, our regret bounds are optimal up to logarithmic gaps in $n$ and $\rho^{-1}$.

\paragraph{Remark.}
Under exact communication, Ascent-FTRL reduces to AD-FTGL \citep{wan2024optimal} with block size $L=K=\Theta(\rho^{-1/2}\ln n)$ and $M=1$. It achieves regret bounds of $O(n\rho^{-1/4}\sqrt{T\ln n})$ and $O(n\rho^{-1/2}\ln n\ln T)$ for convex and strongly convex loss functions, respectively, matching the optimal results for D-OCO with exact communication \citep{wan2024nearly,wan2024optimal}.

\section{Conclusion and Future Work}

This paper investigates D-OCO with compressed communication and closes the gap between the existing regret upper and lower bounds. We first identify the key limitation of existing algorithms and develop Ascent-gossip, which achieves faster consensus under compressed communication. We then incorporate Ascent-gossip into D-OCO and propose Ascent-FTRL, which achieves regret bounds of $\widetilde O(\omega^{-1/2}n\rho^{-1/4}\sqrt{T})$ and $\widetilde O(\omega^{-1}n\rho^{-1/2}\ln T)$ for convex and strongly convex loss functions, respectively.

While this work focuses on D-OCO, Ascent-gossip is a general gossip algorithm for compressed communication and does not rely on the specific structure of online learning. An interesting direction for future work is to apply Ascent-gossip to decentralized offline stochastic optimization and investigate whether it can improve the dependence of existing convergence guarantees on the spectral gap $\rho$.

\bibliography{references}

\newpage
\appendix

\section{Additional Related Work}\label{additional}

A line of research on communication-efficient D-OCO focuses on reducing the frequency of inter-node communication. \citet{zhang2017projection} propose the distributed online conditional gradient algorithm (D-OCG), which replaces projections with linear optimization to reduce the computational cost of local updates, while still requiring neighboring learners to communicate at every round. \citet{wan2020projection} propose D-BOCG, which reduces the number of communication rounds to $O(\sqrt{T})$ while achieving the same $O(T^{3/4})$ regret bound as that of \citet{zhang2017projection} for convex loss functions. Subsequently, \citet{wan2022projection} exploit strong convexity to achieve an $O(T^{2/3}(\ln T)^{1/3})$ regret bound with only $O(T^{1/3}(\ln T)^{2/3})$ communication rounds. They further establish corresponding lower bounds on the communication complexity.

\begin{algorithm}[t]
\floatconts{alg:ascent-matrix}
{\caption{Ascent-gossip in matrix notation}}{%
\begin{algorithmic}[1]
\STATE \textbf{Input:} $X_0$, consistent shared memories $\hat X_0$,
positive integers $K,M$, and $P,\theta$
\STATE Set $X_{-1}=X_0$
\FOR{$t=0,\ldots,K-1$}
\STATE $Q_t=\FCC(X_t-\hat X_t,M)$
\STATE $\hat X_{t+1}=\hat X_t+Q_t$
\STATE $X_{t+1}=(1+\theta)
\bigl(X_t+\hat X_{t+1}(P-I)\bigr)-\theta X_{t-1}$
\ENDFOR
\end{algorithmic}
}
\end{algorithm}

\section{Proof of Theorems}
\subsection{Proof of Theorem~\ref{thm:acg}}
\label{app:acg}

In this section, we present the convergence rate of our proposed Ascent-gossip. We first introduce the matrix notation used in the proof. We define
\begin{equation*}
X_t=\left[\vx_1^{(t)},\ldots,\vx_n^{(t)}\right]
\in\R^{d\times n},
\quad 
Q_t=\left[\q_1^{(t)},\ldots,\q_n^{(t)}\right]
\in\R^{d\times n},
\quad
\hat{X}_t=\left[\hat\vx_1^{(t)},\ldots,\hat\vx_n^{(t)}\right]
\in\R^{d\times n}.
\end{equation*}
Let $J=\mathbf1\mathbf1^\top/n$ and $\bar X=X_0J$. We further have
\begin{equation*}
    A_t = X_t - X_0J, \qquad B_t = X_t - \hat{X}_t, \qquad e_t = \Frob{A_t}^2 + \Frob{B_t}^2.
\end{equation*}
  
Using this notation, Algorithm~\ref{alg:acg} can be written in the matrix form given in Algorithm~\ref{alg:ascent-matrix}, where $\FCC(\cdot,M)$ is applied columnwise. We also have
\begin{equation}\label{aterm}
    A_{t+1}  = (1+\theta)A_tP - \theta A_{t-1} + (1+\theta)(\hat{X}_{t+1} - X_t)(P-I).
\end{equation}
 
 We provide the following lemma to demonstrate that Ascent-gossip preserves the average state over rounds.
\begin{lemma}[Average preservation]
\label{lem:avg-pres}
For every $t\ge0$, Ascent-gossip ensures $X_tJ=X_0J$.
\end{lemma}

For compression error, we have the following guarantee
\begin{equation*}
    \E\left[\Frob{X_t - \hat{X}_{t+1}}^2\right]= \E\left[\Frob{\hat X_t+\FCC(X_t-\hat X_t,M)-X_t}^2\right] \le(1-\omega)^M\E\left[\Frob{X_t - \hat{X}_t}^2\right] \le(1-\omega)^M\E[e_t].
\end{equation*}
Moreover, since $0\preceq P \preceq I$ and $\theta <1$, we can obtain
\begin{equation}\label{eq:1}
    \Frob{(1+\theta)(\hat{X}_{t+1} - X_t)(P-I)} \leq 2 \Frob{\hat{X}_{t+1} - X_t},
\end{equation}
and we have 
\begin{equation}\label{bterm}
    B_t = A_t - A_{t-1} - (\hat{X}_{t} - X_{t-1}).
\end{equation}
It follows from the above relations that $B_t$ is the sum of the
difference between two consecutive consensus errors and the compression
residual. Therefore, to control the joint error $e_t$, we need to bound
both $A_t$ and $A_t-A_{t-1}$. We introduce the following lemma.

\begin{lemma}
\label{lem:ascent-state-bound}
Suppose that $P$ satisfies Assumption~\ref{ass:matrix}.
Let $X_t$ and $\hat X_t$ be generated by Algorithm~\ref{alg:acg}
with $X_{-1}=X_0$ and
$\theta=(1+\sqrt{2\rho-\rho^2})^{-1}$.
Then, for every $t\ge1$, we have 
\begin{align*}
\Frob{A_t}^2+\Frob{A_t-A_{t-1}}^2\le\left(\sqrt{3+\sqrt5}\,r^t\Frob{A_0}+\frac{4\sqrt{4+2\sqrt2}}{\sqrt{3\rho}} \sum_{s=0}^{t-1}r^{t-1-s} \Frob{X_s-\hat X_{s+1}}\right)^2,
\end{align*}
where $A_t=X_t-X_0J$ and $r=1-(1-1/\sqrt2)\sqrt\rho$.
\end{lemma}

Combining \eqref{bterm} with
Lemma~\ref{lem:ascent-state-bound}, we obtain
\begin{align*}
\sqrt{e_t}&\leq \sqrt{\Frob{A_t}^2 + \Frob{A_t -A_{t-1}}^2}+\Frob{X_{t-1} - \hat{X_t}}\\
&\le\sqrt{3+\sqrt5}\,r^t\Frob{A_0}
 +\frac{4\sqrt{4+2\sqrt2}}{\sqrt{3\rho}}
 \sum_{s=0}^{t-1}r^{t-1-s}\Frob{\hat{X}_{s+1} - X_s}+\Frob{\hat{X}_{t} - X_{t-1}}\\
 &\leq \sqrt{3+\sqrt5}\,r^t\Frob{A_0} + \frac{5\sqrt{2}}{\sqrt{\rho}}\sum_{s=0}^{t-1}r^{t-1-s}\Frob{\hat{X}_{s+1} - X_s},
\end{align*}
where the last inequality is due to $4\sqrt{4+2\sqrt2}/\sqrt3+1\le5\sqrt2$. Using Minkowski’s inequality yields
\begin{equation}\label{eq:2}
\begin{aligned}
    \sqrt{\E[e_t]} &\leq \sqrt{3+\sqrt5}\,r^t\sqrt{\E\left[\Frob{A_0}^2\right]} + \frac{5\sqrt{2}}{\sqrt{\rho}}\sum_{s=0}^{t-1}r^{t-1-s}\sqrt{\E\left[\Frob{\hat{X}_{s+1} - X_s}^2\right]}\\
    &\leq  \sqrt{3+\sqrt5}\,r^t\sqrt{\E\left[e_0\right]} + \frac{5\sqrt{2(1-\omega)^M}}{\sqrt{\rho}}\sum_{s=0}^{t-1}r^{t-1-s}\sqrt{\E\left[e_s\right]}\\
    &\leq  \sqrt{3+\sqrt5}\,r^t\sqrt{\E\left[e_0\right]} + \frac{2-\sqrt{2}}{4}\sqrt{\rho}\sum_{s=0}^{t-1}r^{t-1-s}\sqrt{\E\left[e_s\right]},
\end{aligned}    
\end{equation}
where the second inequality is due to $\Frob{A_0}^2\leq e_0$ and the last inequality is due to $(1-\omega)^M \leq \frac{\rho^2}{2401}$ and $\frac{5\sqrt{2(1-\omega)^M}}{\sqrt\rho} \le \frac{5\sqrt2}{49}\sqrt\rho <\frac{2-\sqrt2}{4}\sqrt\rho$.
We now prove by \emph{induction} that
\begin{equation}\label{goal}
    \sqrt{\E\left[e_t\right]}\leq \sqrt{3+\sqrt{5}}\left(1- \frac{2-\sqrt{2}}{4}\sqrt{\rho}\right)^t\sqrt{\E\left[e_0\right]}.
\end{equation}

First, \eqref{goal} holds when $t=0$ because $ \sqrt{3+\sqrt{5}} > 1.$ Assume \ref{goal} holds for $t=1,\dots,k$. When $t=k+1,$ we have the following equality
\begin{equation}\label{eq:3}
\begin{aligned}
&\left(1-\frac{2-\sqrt{2}}{2}\sqrt{\rho}\right)^t+\frac{2-\sqrt{2}}{4}\sqrt{\rho}
\sum_{k=0}^{t-1}\left(1-\frac{2-\sqrt{2}}{2}\sqrt{\rho}\right)^{t-1-k}\left(1-\frac{2-\sqrt{2}}{4}\sqrt{\rho}\right)^k\\
=& \left(1-\frac{2-\sqrt{2}}{2}\sqrt{\rho}\right)^t+\sum_{k=0}^{t-1}
\Biggl[\left(1-\frac{2-\sqrt{2}}{2}\sqrt{\rho}\right)^{t-1-k} \left(1-\frac{2-\sqrt{2}}{4}\sqrt{\rho}\right)^{k+1}\\
&\qquad-\left(1-\frac{2-\sqrt{2}}{2}\sqrt{\rho}\right)^{t-k}\left(1-\frac{2-\sqrt{2}}{4}\sqrt{\rho}\right)^k\Biggr]\\
=&\left(1-\frac{2-\sqrt{2}}{2}\sqrt{\rho}\right)^t+\sum_{k=1}^{t}\left(1-\frac{2-\sqrt{2}}{2}\sqrt{\rho}\right)^{t-k}\left(1-\frac{2-\sqrt{2}}{4}\sqrt{\rho}\right)^k\\
&\qquad-\sum_{k=0}^{t-1}\left(1-\frac{2-\sqrt{2}}{2}\sqrt{\rho}\right)^{t-k}\left(1-\frac{2-\sqrt{2}}{4}\sqrt{\rho}\right)^k\\
=&\left(1-\frac{2-\sqrt{2}}{2}\sqrt{\rho}\right)^t+\left(1-\frac{2-\sqrt{2}}{4}\sqrt{\rho}\right)^t-\left(1-\frac{2-\sqrt{2}}{2}\sqrt{\rho}\right)^t\\
=&\left(1-\frac{2-\sqrt{2}}{4}\sqrt{\rho}\right)^t.
\end{aligned}    
\end{equation}

By combining \eqref{eq:2} and \eqref{eq:3}, we have 
\begin{align*}
        \sqrt{\E[e_{k+1}]}  &\leq  \sqrt{3+\sqrt5}\,r^{k+1}\sqrt{\E\left[e_0\right]} + \frac{2-\sqrt{2}}{4}\sqrt{\rho}\sum_{s=0}^{k}r^{k-s}\sqrt{\E\left[e_s\right]}\\
        &\leq\sqrt{3+\sqrt5}r^{k+1}\sqrt{\E\left[e_0\right]} + \frac{2-\sqrt{2}}{4}\sqrt{\rho}\sum_{s=0}^{k}r^{k-s} \sqrt{3+\sqrt{5}}\left(1- \frac{2-\sqrt{2}}{4}\sqrt{\rho}\right)^s\sqrt{\E\left[e_0\right]}\\
         &=\sqrt{3+\sqrt5}\sqrt{\E\left[e_0\right]}\left(r^{k+1} + \frac{2-\sqrt{2}}{4}\sqrt{\rho}\sum_{s=0}^{k}r^{k-s}\left(1-\frac{2-\sqrt{2}}{4}\sqrt{\rho}\right)^s \right)\\
         &\leq \sqrt{3+\sqrt5}\left(1-\frac{2-\sqrt{2}}{4}\sqrt{\rho}\right)^{k+1}\sqrt{\E\left[e_0\right]},
\end{align*}
which completes the induction.

Therefore, we have
\begin{equation*}
     \E\left[e_t\right]\leq (3+\sqrt{5})\left(1- \frac{2-\sqrt{2}}{4}\sqrt{\rho}\right)^{2t}\E\left[e_0\right]\leq (3+\sqrt{5})\left(1- \frac{\sqrt{\rho}}{4}\right)^{t}\E\left[e_0\right].
\end{equation*}

\subsection{Proof of Theorem \ref{thm:cvx}}
\label{app:cvx}
Before we present the proof for this theorem, we present some definitions.
We divide the total $T$ round into $T/L$ blocks and define $\bar{\vz}(b)=(1/n)\sum_{i=1}^{n}\mathbf{\vz}_i(b)$, $\Bar{\mathbf{g}}(b) = (1/n)\sum_{i=1}^{n} \mathbf{d}_i(b)$ and $\Bar{\mathbf{g}}_i(b) = \mathbf{d}_i(b)$.  For $b \in [T/L]$, we define the virtual sequence as
\begin{equation}
	\label{eq:11}
	\bar{\mathbf{w}}(b)=\argmin_{\mathbf{w} \in \mathcal{K}}
	\left\langle\bar{\vz}(b), \mathbf{w}\right\rangle
	+\frac{1}{\eta}\|\mathbf{w}\|^2,
\end{equation}
which is the decision updated with the average exact gradient over the past $b-1$ blocks. Then, we introduce some useful lemmas.

\begin{lemma}[Lemma 6.6 in \citet{Garber2016}] \label{lem:gh16}
Let $\{\ell_t(\w)\}_{t=1}^T$ be a sequence of functions and 
$\w_t^\star \in \arg\min_{\w \in \K} \sum_{k=1}^t \ell_k(\w)$ for any $t \in [T]$. 
Then, it holds that
\begin{equation*}
    \sum_{t=1}^T \ell_t(\w_t^\star) - \min_{\w \in \K} \sum_{t=1}^T \ell_t(\w) \le 0.
\end{equation*}
\end{lemma}

\begin{lemma}[Lemma 5 in \citet{Duchi2011}] \label{lem:duchi11}
Let $\Pi_{\K}(\mathbf{u}, \eta) = \arg\min_{\w \in \K} \langle \mathbf{u}, \w \rangle + \frac{1}{\eta} \|\w\|^2$.
For any $\mathbf{u}, \mathbf{v} \in \R^d, \eta >0$, we have
\begin{equation*}
\|\Pi_{\K}(\mathbf{u}, \eta) - \Pi_{\K}(\mathbf{v}, \eta)\| \le \frac{\eta}{2} \|\mathbf{u} - \mathbf{v}\|.    
\end{equation*}
\end{lemma}
Let $\ell_1(\mathbf{w})=\langle\mathbf{w},\bar{\mathbf{g}}(1)\rangle+\frac{1}{\eta}\|\mathbf{w}\|^2$ and $\ell_b(\mathbf{w})=\langle\mathbf{w},\bar{\mathbf{g}}(b)\rangle$ for any $b=2,\ldots,T/L$. Combining Lemma~\ref{lem:gh16} with the definition of $\bar{\mathbf{w}}(b)$ in \eqref{eq:11}, for any $\mathbf{w}\in\mathcal{K}$, we have
\begin{equation}\label{eq:12}
\begin{aligned}
    &\sum_{b=1}^{T/L}\langle\bar{\mathbf{w}}(b+1)-\mathbf{w},\bar{\mathbf{g}}(b)\rangle+\frac{\|\bar{\mathbf{w}}(2)\|^2-\|\mathbf{w}\|^2}{\eta}=\sum_{b=1}^{T/L}\left(\ell_b(\bar{\mathbf{w}}(b+1))-\ell_b(\mathbf{w})\right)\leq0.
\end{aligned}
\end{equation}
Then, it is not hard to verify that
\begin{equation}\label{eq:13}
\begin{aligned}
    \sum_{b=1}^{T/L}\langle\bar{\mathbf{w}}(b)-\mathbf{w},\bar{\mathbf{g}}(b)\rangle
    &=\sum_{b=1}^{T/L}\langle\bar{\mathbf{w}}(b+1)-\mathbf{w},\bar{\mathbf{g}}(b)\rangle+\sum_{b=1}^{T/L}\langle\bar{\mathbf{w}}(b)-\bar{\mathbf{w}}(b+1),\bar{\mathbf{g}}(b)\rangle\\
    &\leq\frac{\|\mathbf{w}\|^2-\|\bar{\mathbf{w}}(2)\|^2}{\eta}+\sum_{b=1}^{T/L}\langle\bar{\mathbf{w}}(b)-\bar{\mathbf{w}}(b+1),\bar{\mathbf{g}}(b)\rangle\\
    &\leq\frac{D^2}{\eta}+\sum_{b=1}^{T/L}\|\bar{\mathbf{w}}(b)-\bar{\mathbf{w}}(b+1)\|\|\bar{\mathbf{g}}(b)\|\\
    &\leq\frac{D^2}{\eta}+\frac{\eta TLG^2}{2},
\end{aligned}
\end{equation}
where the third inequality is due to Lemma~\ref{lem:duchi11}.

We define
\begin{equation*}
    \mathbf{w}_i(b)=\operatorname*{argmin}_{\mathbf{w}\in\mathcal{K}}\left\{\langle\mathbf{w},\mathbf{z}_i(b-1)\rangle+\frac{1}{\eta}\|\mathbf{w}\|^2\right\}.
\end{equation*}

Now, we  derive the regret bound of  learner $i$.

\begin{equation}\label{eq:14}
\begin{aligned}
    & f_{t,j}(\mathbf{w}_i(b))-f_{t,j}(\mathbf{w})\\
    \leq& f_{t,j}(\mathbf{w}_j(b))-f_{t,j}(\mathbf{w})+G\|\mathbf{w}_j(b)-\mathbf{w}_i(b)\|_2\\
    \leq&\langle\nabla f_{t,j}(\mathbf{w}_j(b)),\mathbf{w}_j(b)-\mathbf{w}\rangle+G\|\mathbf{w}_j(b)-\bar{\mathbf{w}}(b)+\bar{\mathbf{w}}(b)-\mathbf{w}_i(b)\|\\
    \leq&\langle\nabla f_{t,j}(\mathbf{w}_j(b)),\bar{\mathbf{w}}(b)-\mathbf{w}\rangle+\langle\nabla f_{t,j}(\mathbf{w}_j(b)),\mathbf{w}_j(b)-\bar{\mathbf{w}}(b)\rangle\\
    &\qquad +G\norm{\mathbf{w}_j(b)-\bar{\mathbf{w}}(b)} + G\norm{\bar{\mathbf{w}}(b)-\mathbf{w}_i(b)}\\
    \leq&\langle\nabla f_{t,j}(\mathbf{w}_j(b)),\bar{\mathbf{w}}(b)-\mathbf{w}\rangle+G\|\mathbf{w}_j(b)-\bar{\mathbf{w}}(b)\|\\
    &\qquad +G\norm{\mathbf{w}_j(b)-\bar{\mathbf{w}}(b)} + G\norm{\bar{\mathbf{w}}(b)-\mathbf{w}_i(b)}.
\end{aligned}
\end{equation}
Then we processed  to bound the term $\norm{\mathbf{w}_j(b)-\bar{\mathbf{w}}(b)}$.

For $b\ge2$, Lemma~\ref{lem:duchi11} gives
\begin{equation*}
\begin{aligned}
\E[\norm{\w_i(b)-\bar{\w}(b)}]
&\le \frac{\eta}{2}
\E[\norm{\vz_i(b-1)-\bar{\vz}(b)}]\\
&\le \frac{\eta}{2}
\E[\norm{\vz_i(b-1)-\bar{\vz}(b-1)}]
+\frac{\eta}{2}\E[\norm{\bar{\mathbf{g}}(b-1)}]\\
&\le \frac{\eta LG}{2}+\frac{\eta LG}{2}
=\eta LG,
\end{aligned}
\end{equation*}
where the last inequality follows from Lemma~\ref{lem:consensus}.
For $b=1$, we have $\w_i(1)=\bar{\w}(1)=\vzero$.
Then we give the final bound for learner $i$. By setting $\eta = \frac{D}{G\sqrt{LT}}$, we can derive
\begin{equation}
\label{eq:cvx-decomp}
\begin{aligned}
\E[R(T,i)]
&\le n\E\left[\sum_{b=1}^{B}
\langle\bar{\mathbf{g}}(b),\bar{\w}(b)-\w\rangle\right]
+3n\eta LTG^2\\
&\le \frac{nD^2}{\eta}
+\frac{n\eta TLG^2}{2}
+3n\eta LTG^2\\
&=\frac{nD^2}{\eta}
+\frac{7n\eta TLG^2}{2}\\
&=\frac92 nDG\sqrt{LT} = O\left(\omega^{-1/2}n\rho^{-1/4}
\sqrt{T\ln n\ln(\rho^{-1})}\right).
\end{aligned}
\end{equation}
\subsection{Proof of Theorem \ref{thm:scvx}}
\label{app:scvx}

In this section, we present the proof for strongly convex loss functions. We define the virtual sequence as
\begin{equation}
	\label{eq:21}
	\bar{\mathbf{w}}(b)=\argmin_{\mathbf{w} \in \mathcal{K}}
	\left\langle\bar{\vz}(b), \mathbf{w}\right\rangle
	+\frac{(b-1)L\alpha}{2}\|\mathbf{w}\|^2,
\end{equation}
where $\bar{\mathbf{z}}(b) = \sum_{k=1}^{b-1} \bar{\mathbf{d}}(k)$
and $\bar{\mathbf{d}}(k) = \frac{1}{n}\sum_{i=1}^{n}\mathbf{d}_i(k)$.
In the following, we will bound the regret of any learner $i$ by analyzing
the regret of $\bar{\mathbf{w}}(2), \ldots, \bar{\mathbf{w}}(T/L+1)$
on a sequence of loss functions defined by $\bar{\mathbf{d}}(1), \ldots, \bar{\mathbf{d}}(T/L)$
and the distance $\|\mathbf{w}_i(b)-\bar{\mathbf{w}}(b+1)\|$ for any $b \in [T/L]$. Let $\ell_b(\w)=\inner{\w}{\bar\vd(b)}+\frac{L\alpha}{2}\norm{\w}^2$. It is easy to verify that
\begin{equation}\label{eq:22}
    \sum_{b=1}^{T/L}\ell_b(\bar\w(b+1))-\sum_{b=1}^{T/L}\ell_b(\w)\leq 0.
\end{equation}

The actual decision sequence satisfies 
\begin{equation*}
    \w_i(b)=\argmin_{\w\in\K}\left\{\inner{\w}{\vz_i(b-1)}+\frac{(b-2)L\alpha}{2}\norm{\w}_2^2\right\}.
\end{equation*}
Thus, we have
\begin{equation*}
    \E\left[\norm{\w_i(b)-\bar\w(b-1)}\right]\leq\frac{\E\left[\norm{\vz_i(b-1)-\bar\vz(b-1)}\right]}{(b-2)L\alpha}\leq\frac{G+\alpha D}{(b-2)\alpha},
\end{equation*}
where the first inequality follows from
Lemma~\ref{lem:duchi11} and the second follows from
Lemma~\ref{lem:consensus}.

To bound $\norm{\w_i(b)-\bar\w(b+1)}$, we need to bound $\norm{\bar\w(b-1)-\bar\w(b+1)}$. For any $\vx,\vy\in\K$, we have  
\begin{equation}
\label{eq:23}
\begin{aligned}
    |\ell_b(\vx)-\ell_b(\vy)|&=\left|\inner{\bar\vd(b)+\frac{L\alpha}{2}(\vx+\vy)}{\vx-\vy}\right|\\
    &\leq L(G+2\alpha D)\norm{\vx-\vy}.
\end{aligned}
\end{equation}

Since $\sum_{k=1}^b\ell_k$ is $(bL\alpha)$-strongly convex, we have
\begin{align*}
    \norm{\bar\w(b-1)-\bar\w(b+1)}^2&\leq\frac{2}{bL\alpha}\sum_{k=1}^b\left(\ell_k(\bar\w(b-1))-\ell_k(\bar\w(b+1))\right)\\
    &\leq\frac{2}{bL\alpha}\sum_{k=b-1}^b\left(\ell_k(\bar\w(b-1))-\ell_k(\bar\w(b+1))\right)\\
    &\leq\frac{4(G+2\alpha D)}{b\alpha}\norm{\bar\w(b-1)-\bar\w(b+1)},
\end{align*}
which further implies
\begin{equation*}
    \norm{\bar\w(b-1)-\bar\w(b+1)} \leq \frac{4(G+2\alpha D)}{b\alpha}.
\end{equation*}
Therefore, we have
\begin{equation}
\label{eq:34}
\begin{aligned}
    \E\left[\norm{\w_i(b)-\bar\w(b+1)}\right]&\leq\E\left[\norm{\w_i(b)-\bar\w(b-1)}\right]+\E\left[\norm{\bar\w(b-1)-\bar\w(b+1)}\right]\\
    &\leq\frac{G+\alpha D}{(b-2)\alpha}+\frac{4(G+2\alpha D)}{b\alpha}\leq\frac{5G+9\alpha D}{(b-2)\alpha}
\end{aligned}
\end{equation}
 and $\E\left[\norm{\w_i(b)-\bar\w(b+1)}\right] \leq D$ when $b=1$ or $2$.

 Now, we are ready to derive the regret bound for learner $i$. By using Assumption \ref{ass:scvx}, we have
 \begin{align*}
    &f_{t,j}(\w_i(b))-f_{t,j}(\w)\\
    \leq& f_{t,j}(\w_j(b))-f_{t,j}(\w)+G\norm{\w_j(b)-\w_i(b)}_2\\
    \leq&\inner{\nabla f_{t,j}(\w_j(b))}{\w_j(b)-\w}-\frac\alpha2\norm{\w_j(b)-\w}^2+G\norm{\w_j(b)-\w_i(b)}\\
    \leq&\inner{\nabla f_{t,j}(\w_j(b))}{\bar\w(b+1)-\w}-\frac\alpha2\norm{\w_j(b)-\w}^2+2G\norm{\w_j(b)-\bar\w(b+1)}+G\norm{\w_i(b)-\bar\w(b+1)}.
\end{align*}
Since $ \norm{\w_j(b)-\w}^2\geq2\inner{\w_j(b)}{\bar\w(b+1)-\w}+\norm{\w}^2-\norm{\bar\w(b+1)}^2,$ we can derive
\begin{align*}
f_{t,j}(\w_i(b))-f_{t,j}(\w)\leq&\inner{\nabla f_{t,j}(\w_j(b))-\alpha\w_j(b)}{\bar\w(b+1)-\w}+\frac\alpha2\left(\norm{\bar\w(b+1)}^2-\norm{\w}^2\right)\\
    &\qquad+2G\norm{\w_j(b)-\bar\w(b+1)}+G\norm{\w_i(b)-\bar\w(b+1)}.
\end{align*}
Finally, we have
\begin{align*}
    &\sum_{b=1}^B\sum_{t=(b-1)L+1}^{bL}\sum_{j=1}^n\left(f_{t,j}(\w_i(b))-f_{t,j}(\w)\right)\\
    \leq&\sum_{b=1}^B\sum_{t=(b-1)L+1}^{bL}\sum_{j=1}^n f_{t,j}(\w_j(b))-f_{t,j}(\w)+G\norm{\w_j(b)-\w_i(b)}_2\\
    \leq&\sum_{b=1}^B\sum_{t=(b-1)L+1}^{bL}\sum_{j=1}^n\inner{\nabla f_{t,j}(\w_j(b))}{\bar\w(b+1)-\w}-\frac\alpha2\norm{\w_j(b)-\w}^2\\
    &\qquad+2G\norm{\w_j(b)-\bar\w(b+1)}+G\norm{\w_i(b)-\bar\w(b+1)}\\
    \leq&\sum_{b=1}^B\sum_{t=(b-1)L+1}^{bL}\sum_{j=1}^n\inner{\nabla f_{t,j}(\w_j(b))-\alpha\w_j(b)}{\bar\w(b+1)-\w}+\frac\alpha2\left(\norm{\bar\w(b+1)}^2-\norm{\w}^2\right)\\
    &\qquad+2G\norm{\w_j(b)-\bar\w(b+1)}+G\norm{\w_i(b)-\bar\w(b+1)}\\
    =&n\sum_{b=1}^B\left(\inner{\bar{\vd}(b)}{\bar\w(b+1)-\w}+\frac{L\alpha}{2}\left(\norm{\bar\w(b+1)}^2-\norm{\w}^2\right)\right) \\
    &\qquad +GL\sum_{b=1}^B\left(2\sum_{j=1}^n \norm{\w_j(b)-\bar\w(b+1)}+n\norm{\w_i(b)-\bar\w(b+1)}\right)\\
    \leq & GL\sum_{b=1}^B\left(2\sum_{j=1}^n \norm{\w_j(b)-\bar\w(b+1)}+n\norm{\w_i(b)-\bar\w(b+1)}\right),
\end{align*}
where the last inequality is due to \eqref{eq:22}.

Therefore, by using \eqref{eq:34}, we can derive
\begin{align*}
    \E\left[R(T,i)\right]&\leq 6nLGD + \frac{3nLG(5G+9\alpha D)}{\alpha}\sum_{b=1}^B \frac{1}{b} \leq \frac{33nLG(G+\alpha D)}{\alpha}\ln{T}\\
    &= O(n\omega^{-1}\rho^{-1/2}\ln{T}\ln{n}\ln{\rho^{-1}}).
\end{align*}

\newpage

\section{Proof of Supporting Lemmas}
\subsection{Proof of Lemma \ref{lem:consensus}}
We define $D(b) = [\vd_1(b),...,\vd_n(b)]\in\R^{d\times n}$ and we have 
\begin{align*}
    \norm{\vd_i(b)}\leq L(G+\alpha D), \norm{\bar{\vd}(b)}\leq L(G+\alpha D),\Frob{D(b)}^2\leq nL^2(G+\alpha D)^2.
\end{align*}
At the beginning of block $b$, we initialize the state as 
\begin{equation*}
    \vz_i^0(b)=\vz_i^{K}(b-1)+\vd_i(b-1), \quad \hat{ \vz}_i^0(b)= \hat{\vz}_i^K(b-1).
\end{equation*}
and we can ensure the average state is fixed over rounds, i.e., $\frac{1}{n}\sum_{i=1}^n\vz_i^0(b)=\frac{1}{n}\sum_{i=1}^n\vz_i^k(b)= \bar{\vz}(b) = \bar{\vz}(b-1)+\bar{\vd}(b-1)$.

Then we perform $K$ rounds Ascent-gossip. We define 
 \begin{equation*}
     e_b^k = \frac{1}{n}\sum_{i=1}^n\E\left[\norm{\vz_i^k(b)-\bar{\vz}(b)}^2+\norm{\vz_i^k(b)-\hat{\vz}_i^k(b)}^2\right]
 \end{equation*}
 and we first bound the error at the beginning of block $b$.
 \begin{align*}
\sqrt{e_b^0}&=\left(\frac1n\sum_{i=1}^n\E\left[\norm{\vz_i^K(b-1)-\bar{\vz}(b-1)+\vd_i(b-1)-\bar{\vd}(b-1)}_2^2\right.\right.\\
&\qquad\left.\left.+\norm{\vz_i^K(b-1)-\hat{\vz}_i^K(b-1)+\vd_i(b-1)}_2^2\right]\right)^{1/2}\\
&\le\sqrt{e_{b-1}^K}+\left(\frac1n\sum_{i=1}^n\E\left[\norm{\vd_i(b-1)-\bar{\vd}(b-1)}_2^2+\norm{\vd_i(b-1)}_2^2\right]\right)^{1/2}\\
&\le\sqrt{e_{b-1}^K}+\sqrt2L(G+\alpha D).
 \end{align*}
 \begin{align*}
    \E\left[e_b^{K}\right]=& \frac{1}{n}\sum_{i=1}^n\E\left[\norm{\vz_i^K(b)-\bar{\vz}(b)}^2+\norm{\vz_i^K(b)-\hat{\vz}_i^K(b)}^2\right]\\
    \leq & \left((3+\sqrt{5})\left(1-\frac{\sqrt\rho}{4}\right)^K\right)\frac{1}{n}\sum_{i=1}^n\E\left[\norm{\vz_i^0(b)-\bar{\vz}(b)}^2+\norm{\vz_i^0(b)-\hat{\vz}_i^0(b)}^2\right]\\
    =& \left((3+\sqrt{5})\left(1-\frac{\sqrt\rho}{4}\right)^K\right)\E\left[e_b^{0}\right].
\end{align*}
By setting $K = \left\lceil\frac{4\ln(24n)}{\sqrt{\rho}}\right\rceil$, we can ensure
\begin{equation*}
    (3+\sqrt{5})\left(1-\frac{\sqrt{\rho}}{4}\right)^K \leq (3+\sqrt{5})\exp\left(-\frac{\sqrt{\rho}}{4}K\right) \leq \frac{3+\sqrt{5}}{24n} \leq \frac{2}{9n},
\end{equation*}
which further implies
\begin{equation*}
    \sqrt{e_b^K}\leq\frac{\sqrt{2}}{3\sqrt{n}}\sqrt{e_b^0}\leq\frac13\sqrt{e_{b-1}^K}+\frac{2}{3\sqrt{n}}L(G+\alpha D),
\end{equation*}
where the last inequality follows from $\sqrt{e_b^0}\leq\sqrt{e_{b-1}^K}+\sqrt{2}L(G+\alpha D)$ and $n\geq2$. Since $e_1^K=0$, recursively applying the above inequality gives
\begin{equation*}
    \sqrt{e_b^K}\leq\frac{2L(G+\alpha D)}{3\sqrt{n}}\sum_{j=0}^{b-2}\left(\frac13\right)^j=\frac{L(G+\alpha D)}{\sqrt{n}}\left(1-\frac{1}{3^{b-1}}\right)\leq\frac{L(G+\alpha D)}{\sqrt{n}}.
\end{equation*}
Finally, we have
\begin{equation*}
\begin{aligned}
    \E\left[\norm{\vz_i(b)-\bar{\vz}(b)}_2\right]&\leq\sqrt{\E\left[\norm{\vz_i^K(b)-\bar{\vz}(b)}_2^2\right]}\leq\sqrt{ne_b^K}\leq L(G+\alpha D).
\end{aligned}
\end{equation*}

\subsection{Proof of Lemma \ref{lem:avg-pres}}
Since all copies of a shared memory receive the same increments, Algorithm~\ref{alg:acg} can be written in matrix form as
\begin{equation}
\label{eq:acg-matrix}
\begin{aligned}
Q_t&=\FCC(X_t-\hat X_t, M), \quad \hat X_{t+1}=\hat X_t+Q_t,\\
X_{t+1}&=(1+\theta)\bigl(X_t+\hat X_{t+1}(P-I)\bigr)-\theta X_{t-1},
\end{aligned}
\end{equation}
where $X_{-1}=X_0$. With exact communication, we have $\hat X_{t+1}=X_t$ and~\eqref{eq:acg-matrix} reduces to the accelerated gossip recursion without compression.

Multiplying~\eqref{eq:acg-matrix} by $J$ on the right and using
$(P-I)J=0$ gives
\begin{equation*}
X_{t+1}J=(1+\theta)X_tJ-\theta X_{t-1}J.
\end{equation*}
The initialization $X_{-1}=X_0$ therefore implies $X_tJ=X_0J$ by induction.
\subsection{Proof of Lemma \ref{lem:ascent-state-bound}}
Since $P$ is symmetric, it admits an orthonormal eigenbasis. By Lemma \ref{lem:avg-pres}, we have $A_t\textbf{1}=0$. Let $\vv \in \mathbb{R}^n$ be a unit eigenvector of $P$ orthogonal to $\mathbf{1}$, and  $\lambda \in [0,1-\rho]$ be its eigenvalue, so that $P\vv = \lambda \vv$. We have 
\begin{equation}\label{eq:4}
    A_{t+1}\vv = (1+\theta)\lambda A_t\vv - \theta A_{t-1}\vv + (1+\theta)(1-\lambda)(X_t - \hat{X}_{t+1})\vv.
\end{equation}
We define $q=\sqrt\theta$ and choose $\phi\in(0,\pi/2]$ so that
\begin{equation*}
\cos\phi=\frac{(1+\theta)\lambda}{2\sqrt\theta}
=\frac{(q+q^{-1})\lambda}{2}.
\end{equation*}
This angle is well defined because
\begin{align*}
\sin^2\phi \ge1-\frac{(1+\theta)^2(1-\rho)^2}{4\theta}\notag =\frac{(2\rho-\rho^2)(3+\sqrt{2\rho-\rho^2})}{4}
\ge\frac{3\rho}{4}>0.
\end{align*}

Moreover, since $2\rho-\rho^2\ge\rho$, and $\sqrt{1+\sqrt\rho}\le\sqrt2$, we have 
\begin{align*}
 q &=\frac1{\sqrt{1+\sqrt{2\rho-\rho^2}}} \le\frac1{\sqrt{1+\sqrt\rho}}=1-\frac{\sqrt\rho} {\sqrt{1+\sqrt\rho}\bigl(1+\sqrt{1+\sqrt\rho}\bigr)}\le1-\frac{\sqrt\rho}{\sqrt2(1+\sqrt2)}\\ &=1-\left(1-\frac1{\sqrt2}\right)\sqrt\rho =r. 
 \end{align*}

Let $p_{-1}=p_0=1$ and $p_{t+1}=(1+\theta)\lambda p_t-\theta p_{t-1}$.
Then $p_tA_0\vv$ is the contribution of the initial state to $A_t\vv$ in~\eqref{eq:4}.
The characteristic equation $z^2-(1+\theta)\lambda z+\theta=0$ has roots $q(\cos\phi\pm\mathrm{i}\sin\phi)$. Using the initial conditions, we obtain $p_t=q^t\bigl(\cos(t\phi)+b\sin(t\phi)\bigr), b=\frac{\cos\phi-q}{\sin\phi}.$ We further derive 
\begin{equation*}
p_t-p_{t-1}=-q^t\sin(t\phi)\frac{q+q^{-1}-2\cos\phi}{\sin\phi}.
\end{equation*}
The definition of $\theta$ implies
\begin{equation*}
(q+q^{-1})^2 =4+\frac{2\rho-\rho^2}{1+\sqrt{2\rho-\rho^2}}\le4+2\rho.
\end{equation*}
Hence, we have 
\begin{equation*}
(q+q^{-1})\cos\phi\le\frac{(4+2\rho)(1-\rho)}{2}=2-\rho-\rho^2\le2,
\end{equation*}
which implies
\begin{equation*}
b^2-q^2 =\frac{\cos\phi\bigl((1+q^2)\cos\phi-2q\bigr)}{\sin^2\phi}\le0.
\end{equation*}
Thus $|b|\le q$. Furthermore, for $0\le x\le\cos\phi$,
\begin{equation*}
\frac{d}{dx}\frac{q+q^{-1}-2x}{\sqrt{1-x^2}}=\frac{(q+q^{-1})x-2}{(1-x^2)^{3/2}}\le0.
\end{equation*}
Evaluating this decreasing function at $x=\cos\phi$ and $x=0$ yields $\frac{q+q^{-1}-2\cos\phi}{\sin\phi}\le q+q^{-1}$.
Since $q^2=\theta\in[1/2,1)$, we have 
\begin{equation*}
b^2+\left(\frac{q+q^{-1}-2\cos\phi}{\sin\phi}\right)^2\le q^2+(q+q^{-1})^2=2\theta+2+\theta^{-1}\le5.
\end{equation*}
Consequently, for every $t\ge1$, we have
\begin{equation*}
\begin{aligned}
q^{-2t}\bigl(p_t^2+(p_t-p_{t-1})^2\bigr) &=\cos^2(t\phi)+2b\cos(t\phi)\sin(t\phi) +\left(b^2+\left(\frac{q+q^{-1}-2\cos\phi}{\sin\phi}\right)^2\right)\sin^2(t\phi)\\ &\le\cos^2(t\phi)+2|\cos(t\phi)\sin(t\phi)|
+5\sin^2(t\phi)\\
&\le3+\sqrt5,
\end{aligned}
\end{equation*}
where the last inequality is because that $3+\sqrt5$ is the largest eigenvalue of $\left(\begin{matrix}1&1\\1&5\end{matrix}\right)$. Using $q\le r$, we  obtain
\begin{equation*}
p_t^2+(p_t-p_{t-1})^2\le(3+\sqrt5)r^{2t}.
\end{equation*}
This bound controls the contribution of the initial consensus
error $A_0\vv$ to both $A_t\vv$ and $(A_t-A_{t-1})\vv$.
To bound the full iterates, we must also account for the terms $(1+\theta)(1-\lambda)(X_s-\hat X_{s+1})\vv$ in~\eqref{eq:4}. We next estimate how each of these terms propagates through subsequent iterations.

Fix a realization of the compression randomness and regard the residuals as given in~\eqref{eq:4}. For any $s\ge0$, the term $(1+\theta)(1-\lambda)(X_s-\hat X_{s+1})\vv$ first enters $A_{s+1}\vv$ with coefficient $1$. Its coefficients at subsequent iterations satisfy
the same homogeneous recurrence. We define  $h_{-1}=0$, $h_0=1$, and $h_{\ell+1}=(1+\theta)\lambda h_\ell-\theta h_{\ell-1}$. By linearity of the recurrence, the contribution of the term
from iteration $s$ to $A_t\vv$, for $t\ge s+1$, is
\begin{equation*}
(1+\theta)(1-\lambda)
h_{t-1-s}(X_s-\hat X_{s+1})\vv.
\end{equation*}
Thus, it remains to bound $h_\ell$ and $h_\ell-h_{\ell-1}$
jointly. To obtain these bounds, we first solve the recurrence
\begin{equation*}
h_{\ell+1}=(1+\theta)\lambda h_\ell-\theta h_{\ell-1},
\end{equation*}
with the initial conditions $h_{-1}=0$ and $h_0=1$.
Its characteristic equation is
\begin{equation*}
z^2-(1+\theta)\lambda z+\theta=0.
\end{equation*}
Using $q=\sqrt\theta$ and $\cos\phi=(1+\theta)\lambda/(2q)$, the two roots are
\begin{equation*}
\begin{aligned}
z=\frac{(1+\theta)\lambda\pm\mathrm{i}\sqrt{4\theta-(1+\theta)^2\lambda^2}}{2}=q\bigl(\cos\phi\pm\mathrm{i}\sin\phi\bigr).
\end{aligned}
\end{equation*}
The recurrence and the initial conditions imply
\begin{equation*}
h_0=1,\qquad h_1=(1+\theta)\lambda h_0-\theta h_{-1}=(1+\theta)\lambda=2q\cos\phi.
\end{equation*}
The solution is
\begin{equation*}
\begin{aligned}
h_\ell
&=q^\ell\left(h_0\cos(\ell\phi)+\frac{h_1/q-h_0\cos\phi}{\sin\phi}\sin(\ell\phi)\right)\\
&=q^\ell\left(\cos(\ell\phi)+\frac{\cos\phi}{\sin\phi}\sin(\ell\phi)\right)\\
&=q^\ell\frac{\sin\phi\cos(\ell\phi)+\cos\phi\sin(\ell\phi)}{\sin\phi}=q^\ell\frac{\sin((\ell+1)\phi)}{\sin\phi}.
\end{aligned}
\end{equation*}
We now use this expression to bound $h_\ell$ and $h_\ell-h_{\ell-1}$.
Since $\sin^2\phi\ge3\rho/4$ and $q\le r$, we have
\begin{equation*}
|h_\ell|=q^\ell\frac{|\sin((\ell+1)\phi)|}{\sin\phi}\le
\frac{q^\ell}{\sin\phi}\le\frac{2r^\ell}{\sqrt{3\rho}},
\end{equation*}
which further implies
\begin{equation*}
|h_\ell-h_{\ell-1}|\le |h_\ell|+|h_{\ell-1}|\leq\frac{2(r^\ell+r^{\ell-1})}{\sqrt{3\rho}}.
\end{equation*}

Since $r\ge1/\sqrt2$, we have  
\begin{equation}\label{eq:5}
\sqrt{h_\ell^2+(h_\ell-h_{\ell-1})^2}\le\frac{2\sqrt{4+2\sqrt2}}{\sqrt{3\rho}}r^\ell,
\end{equation}
where the inequality is due to $1+(1+r^{-1})^2\le1+(1+\sqrt2)^2=4+2\sqrt2.$

We now combine the bounds for the initial consensus error and the compression residuals. For the fixed realization of the compression randomness, the recurrence
\begin{equation*}
A_{t+1}\vv=(1+\theta)\lambda A_t\vv-\theta A_{t-1}\vv+(1+\theta)(1-\lambda)(X_t-\hat X_{t+1})\vv
\end{equation*}
and the initial condition $A_{-1}\vv=A_0\vv$ imply
\begin{equation*}
A_t\vv=p_tA_0\vv+(1+\theta)(1-\lambda)\sum_{s=0}^{t-1}h_{t-1-s}(X_s-\hat X_{s+1})\vv.
\end{equation*}
The first term is the contribution of the initial consensus error. Each term in the sum is the contribution of a compression residual introduced at iteration $s$ and propagated through the subsequent $t-1-s$ iterations.

Taking the difference between the expressions for $A_t\vv$ and $A_{t-1}\vv$ yields
\begin{equation*}
(A_t-A_{t-1})\vv=(p_t-p_{t-1})A_0\vv+(1+\theta)(1-\lambda)\sum_{s=0}^{t-1}(h_{t-1-s}-h_{t-2-s})(X_s-\hat X_{s+1})\vv.
\end{equation*}
We further have
\begin{equation*}
\begin{aligned}
&\sqrt{\|A_t\vv\|_2^2+\|(A_t-A_{t-1})\vv\|_2^2}\\
\leq &\sqrt{p_t^2+(p_t-p_{t-1})^2}\,\|A_0\vv\|_2+(1+\theta)(1-\lambda)\sum_{s=0}^{t-1}\sqrt{h_{t-1-s}^2+(h_{t-1-s}-h_{t-2-s})^2}\,\|(X_s-\hat X_{s+1})\vv\|_2.
\end{aligned}
\end{equation*}
By using $p_t^2+(p_t-p_{t-1})^2\le(3+\sqrt5)r^{2t}$ and~\eqref{eq:5}, we have
\begin{equation*}
\sqrt{\|A_t\vv\|_2^2+\|(A_t-A_{t-1})\vv\|_2^2}\le\sqrt{3+\sqrt5}r^t\|A_0\vv\|_2 +\frac{2\sqrt{4+2\sqrt2}}{\sqrt{3\rho}}\sum_{s=0}^{t-1}r^{t-1-s}(1+\theta)(1-\lambda)\|(X_s-\hat X_{s+1})\vv\|_2. 
\end{equation*}

It remains to combine the bounds over all network directions. Let $\vv_1=\mathbf1/\sqrt n,\vv_2,\ldots,\vv_n$ be an orthonormal eigenbasis of $P$, with $P\vv_j=\lambda_j\vv_j$. Since $A_t\mathbf1=A_{t-1}\mathbf1=0$, we have
\begin{equation*}
\Frob{A_t}^2+\Frob{A_t-A_{t-1}}^2=\sum_{j=2}^n\left(\|A_t\vv_j\|_2^2+\|(A_t-A_{t-1})\vv_j\|_2^2\right).
\end{equation*}
The preceding bounds hold for every $\lambda_j\in[0,1-\rho]$, $j\ge2$. We further obtain
\begin{equation*}
\begin{aligned}
&\sqrt{\Frob{A_t}^2+\Frob{A_t-A_{t-1}}^2}\\
\leq&\sqrt{3+\sqrt5}\,r^t\left(\sum_{j=2}^n\|A_0\vv_j\|_2^2\right)^{1/2}+\frac{2\sqrt{4+2\sqrt2}}{\sqrt{3\rho}}\sum_{s=0}^{t-1}r^{t-1-s}\left(\sum_{j=2}^n(1+\theta)^2(1-\lambda_j)^2\|(X_s-\hat X_{s+1})\vv_j\|_2^2\right)^{1/2}.
\end{aligned}
\end{equation*}
By orthonormality and $A_0\mathbf1=0$, we have  $\sum_{j=2}^n\|A_0\vv_j\|_2^2=\Frob{A_0}^2.$ Moreover, $(I-P)\vv_j=(1-\lambda_j)\vv_j$ and $(I-P)\vv_1=0$, so
\begin{equation*}
\sum_{j=2}^n(1+\theta)^2(1-\lambda_j)^2\|(X_s-\hat X_{s+1})\vv_j\|_2^2 =\Frob{(1+\theta)(X_s-\hat X_{s+1})(I-P)}^2.
\end{equation*}
Therefore, we have 
\begin{equation}\label{eq:6}
\sqrt{\Frob{A_t}^2+\Frob{A_t-A_{t-1}}^2}\le\sqrt{3+\sqrt5}\,r^t\Frob{A_0}+\frac{2\sqrt{4+2\sqrt2}}{\sqrt{3\rho}}\sum_{s=0}^{t-1}r^{t-1-s}\Frob{(1+\theta)(X_s-\hat X_{s+1})(I-P)}.
\end{equation}

Finally, we have 
\begin{equation}\label{eq:7}
\Frob{(1+\theta)(X_s-\hat X_{s+1})(I-P)}\le(1+\theta)\Frob{X_s-\hat X_{s+1}}\|I-P\|_2
\le2\Frob{X_s-\hat X_{s+1}},
\end{equation}
where the second inequality is due to  $\|I-P\|_2\le1$ and $\theta <1.$

Substituting \eqref{eq:7} into \eqref{eq:6} completes the proof.

\end{document}